\documentclass[final]{cvpr}

\usepackage{times}
\usepackage{epsfig}
\usepackage{graphicx}
\usepackage{amsmath}
\usepackage{amssymb}
\usepackage{mmstyle}
\usepackage{multirow}
\usepackage{mathrsfs}
\usepackage[table,xcdraw]{xcolor}
\usepackage{bbm}
\usepackage{dsfont}
\usepackage{booktabs}
\usepackage{enumitem}

\usepackage[pagebackref=true,breaklinks=true,colorlinks,bookmarks=false]{hyperref}

\def\cvprPaperID{3633} 
\def\confYear{CVPR 2021}

\begin{document}

\title{Adversarial Robustness under Long-Tailed Distribution}

\author{Tong Wu$^{1,5}$,\ \ 
Ziwei Liu$^2$,\ \ 
Qingqiu Huang$^3$,\ \ 
Yu Wang$^4$,\ \ 
Dahua Lin$^{1,5,6,7}$\\

$^1$The Chinese University of Hong Kong,
$^2$S-Lab, Nanyang Technological University,
$^3$Huawei,\\
$^4$Tsinghua University, 
$^5$SenseTime-CUHK Joint Lab, 
$^6$Centre of Perceptual and Interactive Intelligence \\
$^7$Shanghai AI Laboratory \\
\tt\small
\{wt020,dhlin,hq016\}@ie.cuhk.edu.hk, 
ziwei.liu@ntu.edu.sg,
yu-wang@mail.tsinghua.edu.cn\\
}


\maketitle


\begin{abstract}


Adversarial robustness has attracted extensive studies recently by revealing the vulnerability and intrinsic characteristics of deep networks.
However, existing works on adversarial robustness mainly focus on balanced datasets, while real-world data usually exhibits a long-tailed distribution.
To push adversarial robustness towards more realistic scenarios, in this work we investigate the adversarial vulnerability as well as defense under long-tailed distributions.
In particular, we first reveal the negative impacts induced by imbalanced data on both recognition performance and adversarial robustness, uncovering the intrinsic challenges of this problem.
We then perform a systematic study on existing long-tailed recognition methods in conjunction with the adversarial training framework.
Several valuable observations are obtained: 1) natural accuracy is relatively easy to improve, 2) fake gain of robust accuracy exists under unreliable evaluation, and 3) boundary error limits the promotion of robustness.
Inspired by these observations, we propose a clean yet effective framework, 
RoBal, which consists of two dedicated modules, a scale-invariant classifier and data re-balancing via both margin engineering at training stage and boundary adjustment during inference.
Extensive experiments demonstrate the superiority of our approach over other state-of-the-art defense methods.
To our best knowledge, we are the first to tackle adversarial robustness under long-tailed distributions, which we believe would be a significant step towards real-world robustness. Our code is available at: \url{https://github.com/wutong16/Adversarial_Long-Tail}.

\end{abstract}
\thispagestyle{empty}


\section{Introduction}
\label{sec:introduction}
Despite the great progress on a variety of computer vision tasks, deep neural networks are found to be vulnerable to minor adversarial perturbations~\cite{szegedy2013intriguing}, \ie, easily misled to make incorrect predictions. The existence of adversarial examples reveals a non-negligible security risk to modern computer vision models, with extensive efforts devoted to improving adversarial robustness.

\begin{figure}[t]
	\centering
	\includegraphics[width=1.0\linewidth]{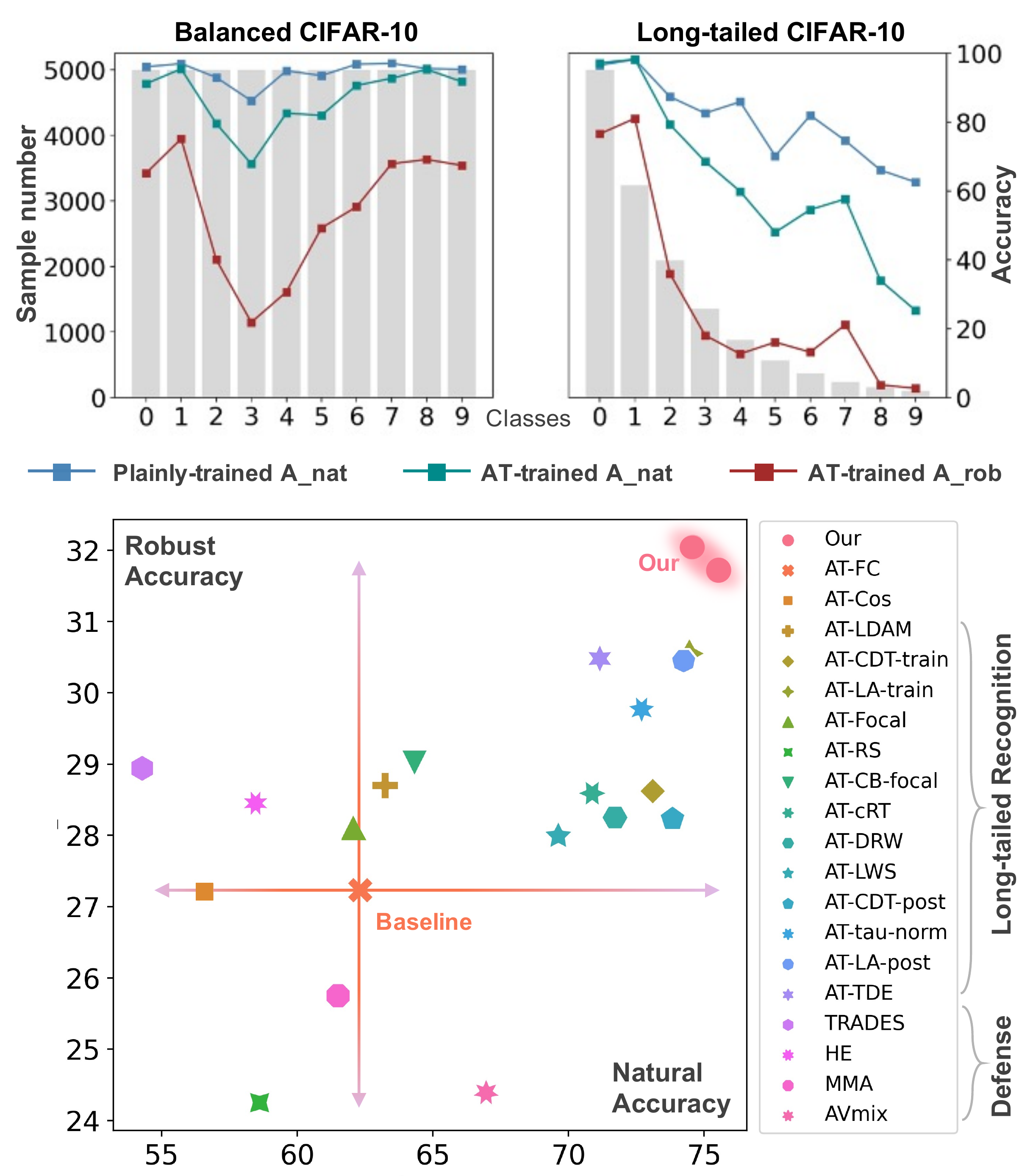}
	\vspace{-10pt}
	\caption{\small
	\textbf{Upper}: A long-tailed data distribution induces decreasing natural and robust accuracy from head to tail and a magnified “sacrifice" of natural accuracy especially to tail classes when adversarial training is applied.
	\textbf{Lower}: Evaluation results on two metric dimensions, including a number of long-tailed recognition methods combined with adversarial training, several state-of-the-art defense methods, and our RoBal in a region with trade-off.
	}
	\vspace{-10pt}
	\label{fig:motivation}
\end{figure}

Existing adversarial robustness research mainly focuses on balanced datasets such as CIFAR and ImageNet~\cite{he2009learning}.
Nevertheless, real-world data usually exhibit a long-tailed distribution~\cite{van2018inaturalist,gupta2019lvis}, which brings challenges not only to the recognition tasks themselves but also to robustness against adversarial attacks.
The former has been attracting increasing attention recently, with a number of algorithms \cite{liu2019largescale, kang2019decoupling, zhou2020bbn, cui2019cb, cao2019ldam, wu2020distribution, wang2021seesaw} proposed to tackle the issue; On the other hand, the latter remains largely unexplored.

To cast light on the challenges of adversarial robustness in \textbf{long-tailed recognition (LT)}, we first perform an intuitive comparison between networks trained on the balanced and long-tailed versions of CIFAR-10, respectively.
Apart from normally trained models, we also adopt the \textbf{adversarial training (AT)} framework~\cite{madry2018towards}, which is one of the most effective and widely used defense methods, to provide the basic adversarial robustness for the networks. 
Per-class classification recalls are evaluated on clean images and images permuted by PGD attack~\cite{madry2018towards}, denoted by \textit{natural accuracy} $\mathbf{A}_{nat}$ and \textit{robust accuracy} $\mathbf{A}_{rob}$, respectively.
$\mathbf{A}_{nat}$ is evaluated on both \textbf{plain} models and \textbf{AT-trained} models, while $\mathbf{A}_{rob}$ is performed only on the latter. Results are visualized in Fig.~\ref{fig:motivation}.
There are three main observations from the comparison: 1) $\mathbf{A}_{nat}$ on plain models drops from head to tail, which is exactly what traditional long-tailed recognition aims to solve. 2) A similar decreasing tendency reasonably occurs in $\mathbf{A}_{rob}$. 3) It is worth noting that 
$\mathbf{A}_{nat}$ drops more significantly at the tail when adversarial training is applied, indicating that the well-known “sacrifice” of the natural accuracy induced by adversarial training is further magnified for tail classes under a long-tailed distribution.

To form a better understanding of the problem, the relationship between natural and robust accuracy can be connected by \textit{boundary error} $\mathbf{R}_{bdy}$~\cite{zhang2019trades} as:
\vspace{-3pt}
\begin{equation}
    \mathbf{A}_{rob} = \mathbf{A}_{nat} - \mathbf{R}_{bdy},
    \label{eq:error_decomposition}
\end{equation}
\vspace{-3pt}
where $\mathbf{R}_{bdy}$ represents how likely the features of clean and correctly predicted inputs are close to the $\epsilon$-extension of the decision boundary. It represents the gap between the two forms of accuracy and indicates the vulnerability of samples against adversarial attacks.

Hence, to achieve improvement on both recognition performance and adversarial robustness, a natural idea is to raise $\mathbf{A}_{nat}$ while keeping a small value of $\mathbf{R}_{bdy}$.
Specifically, on the one hand, we are able to address the issue of imbalance in data distribution via re-balancing strategies, thus we conduct a systematic study of currently widely used long-tailed recognition approaches to explore the proper combinations of these methods and the adversarial training framework. 
On the other hand, we would analyze why a normalized embedding space promotes model resistance against attacks, and then a scale-invariant classifier is introduced to replace the final linear layer.
The idea of data re-balancing is then well aligned with the cosine classifier by the cooperation of class-aware and pair-aware margins during training and boundary adjustment at inference.

Note that the imbalance in data distribution and the deficiency in sample numbers are two issues induced simultaneously when we turn to long-tailed datasets instead of the artificially balanced ones.
Although the importance of data scale in adversarial robustness has been widely studied~\cite{schmidt2018moredata}, we mainly focus on the problem of imbalance in this paper. We study the effect of them separately in Sec~\ref{sec:experiment}, verifying that eliminating prediction priors is crucial to reducing the vulnerability of tail classes under attack.

Our contributions are as follows:
\textbf{1)} To our best knowledge, we are the first to tackle adversarial robustness under long-tailed distribution, which we believe would be a significant step towards real-world robustness. 
\textbf{2)} We conduct a systematic study on existing long-tailed recognition methods and their adoption into the adversarial training procedure. Important insights are gained based on experimental observations.
\textbf{3)} We further develop a clean yet effective approach, RoBal, that achieves state-of-the-art performance on both natural and robust accuracy.



\section{Related Works}
\label{sec:related}

\noindent\textbf{Long-Tailed Recognition.}
To tackle the long-tailed recognition problem, traditional re-balancing approaches include re-sampling~\cite{chawla2002smote, japkowicz2002class, he2009learning, shen2016relay, buda2018systematic} and re-weighting~\cite{cui2019cb, buda2018systematic}. However, these methods may suffer from the issue of under-representing major classes and over-fitting minor ones.
To mitigate these negative impacts, more flexible usages of the basic methods were proposed, such as decoupled training~\cite{kang2019decoupling, zhou2020bbn} and deferred re-balancing schedule~\cite{cao2019ldam}, respectively, and they are proved to be more effective. 
Further, recently proposed approaches address class-specific properties by perspectives like margin~\cite{cao2019ldam}, bias~\cite{ren2020balanced-softmax,menon2020logit}, temperature~\cite{xie2019intriguing} or weight scale~\cite{kang2019decoupling, kim2020adjusting}, and some of these methods can be either adopted to the whole training process or in a post-processing manner.
Another trend of works focuses on sample-specific properties via hard example mining~\cite{lin2017focal} or sample-aware re-weighting strategies leveraging meta-learning~\cite{ren2018l2r, jamal2020rethinking, han2018coteaching}.
Besides, several recent approaches propose to transfer knowledge from head to tail through memory module~\cite{liu2019largescale}, inter-class feature transferring~\cite{liu2020deep}, and “major-to-minor” translation~\cite{kim2020m2m}.
In this paper, we would revisit and summarize a number of these methods and explore their effective combination with adversarial training in Sec.~\ref{sec:empirical_study}.
\begin{table*}[htbp]
  \centering
  \caption{A systematic study of current LT strategies combined with AT framework, detailed explanations are to be included in 
  Sec.~\ref{supp:subsec:implementation_detail_LT}.
  \textcolor[rgb]{ .439, .678, .278}{Green}, \textcolor[rgb]{ .753,  0,  0}{red}, and \textcolor[rgb]{ .267,  .447,  .769}{blue} denote impressive $A_{nat}$, unreliable evaluation of $A_{rob}$ under PGD attack, and the smallest $R_{bdy}$, respectively. 
  }
  \small
    \begin{tabular}{c|l|l|cccc}
    \toprule
    \toprule
    \multicolumn{1}{p{2em}|}{\textbf{Stage}} & \multicolumn{1}{c|}{\textbf{Methods}} & \multicolumn{1}{c|}{\textbf{Formulation}} & \multicolumn{1}{p{2em}}{\textbf{Clean}} & \multicolumn{1}{p{2em}}{\textbf{PGD}} & \multicolumn{1}{p{2em}}{\textbf{AA}} & \multicolumn{1}{p{2em}}{\textbf{Gap}} \\
    \midrule
    \multirow{9}[18]{*}{Train} & Vanilla FC & $g_i = W_i^Tf(x)$ & 62.33  & 29.30  & 28.15  & 34.18  \\
\cmidrule{2-7}          & Vanilla Cos & $g_i = \widetilde W_i^T \widetilde f(x)$ & 56.59  & 29.38  & 27.23  & \textcolor[rgb]{ .267,  .447,  .769}{29.36 } \\
\cmidrule{2-7}          & Class-aware margin~\cite{cao2019ldam} & $g_i = W_i^Tf(x) - \mathbbm{1}\{i=y\} \cdot \delta_i$ & 63.24  & 29.81  & 28.70  & 34.54  \\
\cmidrule{2-7}          & Cosine with margin~\cite{wang2018cosface} & $g_i = \widetilde W_i^T \widetilde f(x) - \mathbbm{1}\{i=y\} \cdot m$ & 58.47  & 31.73  & 28.45  & \textcolor[rgb]{ .267,  .447,  .769}{30.02}  \\
\cmidrule{2-7}          & Class-aware temperature~\cite{ye2020identifying} & $g_i = W_i^Tf(x)\cdot (n_i / n_{max})^\gamma$ & \textcolor[rgb]{ .573,  .816,  .314}{73.11 } & 30.12  & 28.62  & 44.49  \\
\cmidrule{2-7}          & Class-aware bias~\cite{menon2020logit, ren2020balanced-softmax} & $g_i = W_i^Tf(x) + \tau \log(n_i)$ & \textcolor[rgb]{ .573,  .816,  .314}{74.46 } & 32.45  & 30.55  & 43.91  \\
\cmidrule{2-7}          & Hard-exmaple mining~\cite{lin2017focal} & $r(y) = \left( 1-p_y \right)^\gamma$, applyed with BCE loss & 62.07  & 30.73  & 28.12  & 33.95  \\
\cmidrule{2-7}          & Re-sampling~\cite{shen2016relay} & $r_s(i) \propto 1/n_i$    & 58.62  & 25.06  & 24.25  & 34.37  \\
\cmidrule{2-7}          & Re-weighting~\cite{cui2019cb} & $r(y) = (1-\beta) / (1 - \beta^n_y)$ & 64.33  & 34.53  & 29.01  & 35.32  \\
    \midrule
    \multicolumn{1}{c|}{\multirow{3}[6]{*}{Fine-tune}} & One-epoch re-sampling~\cite{kang2019decoupling} & $h_i = {W'}_i^Tf(x)$, ${W'}_i$ re-trained with RS & \textcolor[rgb]{ .573,  .816,  .314}{70.88 } & 29.81  & 28.59  & 42.29  \\
\cmidrule{2-7}          & One-epoch re-weighting~\cite{cao2019ldam, cui2019cb} & $h_i = {W'}_i^T f(x)$, ${W'}_i$ fine-tuned with RW & \textcolor[rgb]{ .573,  .816,  .314}{71.72 } & 32.34  & 28.25  & 43.47  \\
\cmidrule{2-7}          & Learnable classifier scale~\cite{kang2019decoupling} & $h_i = s_i \cdot W_i^Tf(x)$, where $s_i$ is learnable & \textcolor[rgb]{ .573,  .816,  .314}{69.63 } & 28.81  & 27.99  & 41.64  \\
    \midrule
    \multicolumn{1}{c|}{\multirow{3}[8]{*}{Inference}} & Classifier re-scaling~\cite{ye2020identifying, kim2020adjusting} & $h_i = (W_i / n_i^\tau)^Tf(x)$ & \textcolor[rgb]{ .573,  .816,  .314}{73.84 } & \textcolor[rgb]{ .753,  0,  0}{39.05 } & 28.23  & 45.61  \\
\cmidrule{2-7}          & Classifier normalization~\cite{kang2019decoupling} & $h_i = (W_i / \left\Vert W_i \right\Vert^\tau)^Tf(x)$ & \textcolor[rgb]{ .573,  .816,  .314}{72.70 } & \textcolor[rgb]{ .753,  0,  0}{36.57 } & 29.77  & 42.93  \\
\cmidrule{2-7}          & Class-aware bias~\cite{menon2020logit} & $h_i = W_i^Tf(x) - \tau \log(n_i)$ & \textcolor[rgb]{ .573,  .816,  .314}{74.25 } & 31.95  & 30.45  & 43.80  \\
\cmidrule{2-7}          & Feature disentangling~\cite{tang2020longtailed} & $h_i = W_i^T (f(x)-\alpha \cos(f(x),d)\cdot d)$ & \textcolor[rgb]{ .573,  .816,  .314}{71.16 } & 32.69  & 30.48  & 40.68  \\
    \bottomrule
    \bottomrule
    \end{tabular}%
  \label{tab:long_tail}%
  \vspace{-10pt}
\end{table*}%

\noindent\textbf{Adversarial Robustness.}
Plenty of adversarial defense methods have been proposed to tackle the problem of adversarial vulnerability.
Among them, adversarial training~\cite{madry2018towards} is one of the most effective and reliable strategies. Improvements have been made based on the AT framework via theoretical analysis and loss function examination~\cite{zhang2019trades, Wang2020mart,ding2020mma}. 
Many efforts have also been devoted to exploring different training mechanisms such as metric learning~\cite{mao2019metric}, self-supervised learning~\cite{hendrycks2019using}, and semi-supervised learning~\cite{carmon2019unlabeled}.
Since AT is of the high computational cost and time consumption, another line of works~\cite{shafahi2019free, wong2019fast, zhang2019yopo} was proposed to accelerate the training procedure.
Besides, some general strategies were revealed to be critical to the robustness performance such as label smoothing~\cite{shafahi2019label}, early stop~\cite{rice2020overfitting}, different activation functions~\cite{xie2020smooth}, batch normalization~\cite{xie2020improve}, and embedding space~\cite{pang2020boosting}.
Most recently, Pang~\etal~\cite{pang2020bag} and Gowal~\etal~\cite{gowal2020uncovering} made systematic studies on the effect of basic training settings and some other choices, including model size, data, loss, and activation functions, respectively.
Our method is also built on the AT framework, while we focus on the long-tailed training data distribution to explore how it affects the accuracy and ways for improvement.

\vspace{-3pt}
\section{Long-tailed Recognition with Defense}
\label{sec:empirical_study}

In this section, we first briefly introduce the adversarial training (AT) framework.
Then we conduct a systematic study on some popular long-tailed recognition (LT) strategies and explore their proper combination with the AT framework, where the effectiveness is evaluated by $\mathbf{A}_{nat}$, $\mathbf{A}_{rob}$, and $\mathbf{R}_{bdy}$. 
We further reveal a fake increase of robustness that could be induced under unreliable evaluation.
Finally, valuable knowledge from the study is summarized and inspires us to develop our method in Sec.~\ref{sec:method}.

\subsection{Adversarial Training Preliminaries}
Adversarial training, as one of the most effective defense methods, is adopted as the basic framework to maintain basic robustness in this paper. The standard AT and its variants can be formulated as a mini-max problem:
\begin{equation}
\begin{aligned}
           & \min_{\theta}\mathbb{E}_{(x,y) \sim \mathcal{D}}\left[ \mathcal{L}_T (\theta; x + \delta,y) \right], \\ 
  where\ \ &  \delta = \argmax_{\delta \in \mathcal{B}(\epsilon)}\mathcal{L}_A(\theta; x + \delta, y ).
\end{aligned}
\vspace{-3pt}
\label{eq:AT}
\end{equation}
%
The inner optimization aims to find effective adversarial examples by maximizing $\mathcal{L}_A$, and the outer optimization updates network parameters to minimize the training loss $\mathcal{L}_T$.
The standard AT proposed by Madry~\etal~\cite{madry2018towards} uses Cross-Entropy loss(CE) for both $\mathcal{L}_A$ and $\mathcal{L}_T$, while we would explore the effects of different choices in this paper.
\subsection{Revisiting Long-tailed Recognition Methods}
\noindent\textbf{Preliminaries.}
The LT methods, who could be naturally combined with the AT framework, can be categorized into three phases based on different applying stages: \textbf{training}, \textbf{fine-tuning}, and \textbf{inference}, as summarized in Table~\ref{tab:long_tail}. The evaluation metrics include the accuracy of the clean images and permuted images under PGD-20~\cite{madry2018towards} and Auto-Attack (AA)~\cite{croce2020AA}. Details are to be introduced in Sec.~\ref{sec:experiment}.
We also report the gap between clean accuracy and Auto-Attack accuracy for a better view of boundary error.

\noindent\textbf{Notifications.} Suppose there are $C$ classes in total with $n_i, i \in \{1,2,..., C\}$ samples for class $i$. We denote $f(x)$ as the deep feature extracted from image $x$ and $W = [W_1, ..., W_C]$ as the classifier weight vectors. And the normalized weight vectors and features are denoted as $\widetilde W_i =  W_i / \left\Vert W_i \right\Vert$ and $\widetilde f(x) = f / \left\Vert f \right\Vert$.

\noindent\textbf{Training Stage.}
Methods applied to training stage include class-aware \textbf{re-sampling}~\cite{shen2016relay, liu2019largescale, kang2019decoupling, zhou2020bbn}, and several \textbf{cost-sensitive learning} approaches.
We denote the sampling frequency for class $i$ as $r_s(i)$ in Table~\ref{tab:long_tail}.
Cost-sensitive learning methods usually modify the loss function by introducing class-specific parameters like \textit{weight} (CB~\cite{cui2019cb}), ~\textit{margin} (LDAM~\cite{cao2019ldam}),~\textit{bias} (LA-train~\cite{menon2020logit}, Balanced Softmax~\cite{ren2020balanced-softmax}), and \textit{temperature} (CDT-train~\cite{xie2019intriguing}). A~\textit{hard example examining} method (Focal~\cite{lin2017focal}) is also included here although it is applied with binary cross entropy loss.
A general loss function for the cost-sensitive methods above with CE loss can be formulated as:
\vspace{-3pt}
\begin{equation}
\begin{aligned}
    \mathcal{L'}_{CE}(W; f(x),y) & = - r_w(y) \cdot log(\frac{e^{z_y}}{\sum_ie^{z_i}}), \\
    where \ \ z_i & = g_i(W_i, f(x)),
\end{aligned}
\label{eq:loss_ce}
\end{equation}
\vspace{-3pt}
where $g(W, f(x))$ denotes the logit before softmax, and $r_w(y)$ is a re-weighting factor for class $y$.
The widely used linear classifier would have $g_i(W_i, f(x)) = W_i^T f(x) + b_i$.
Different methods on training would have different $g$ functions, which are listed in Table~\ref{tab:long_tail} ( $b_i$ is omit for simplicity).

$\mathcal{L'}_{CE}$ can be adopted to AT procedure in three modes: replacing the CE in $\mathcal{L}_A$, $\mathcal{L}_T$, or both of them, where $\mathcal{L}_A$ and $\mathcal{L}_T$ would affect the optimization of the adversarial examples and network parameter updating, respectively. We empirically observed that modifying $\mathcal{L}_T$ has a more conspicuous influence on the results. Thus results reported here are conducted with the second mode. We present a more detailed study in the supplementary material Sec.~\ref{supp:sec:extensive}.

\noindent\textbf{Fine-tuning Stage.}
Fine-tuning based methods propose to re-train~\cite{kang2019decoupling} or fine-tune the classifier via data re-balancing techniques with the backbone frozen, which take advantage of the idea of decoupling the learning of representation and classifier. We empirically find out that one-epoch of fine-tuning with class-aware sampling or re-weighting would remarkably raise $A_{nat}$ while more steps make little difference.
A similar conclusion is drawn when only weight scales $s_i$ are learned at this stage (LWS~\cite{kang2019decoupling}).

\noindent\textbf{Inference Stage.}
LT methods applied at the inference stage based on a vanilla trained model would usually conduct a different forwarding process from the training stage to address shifted data distributions from train-set to test-set.
Specifically, we denote $h(W, f(x))$ as the logit producing function on inference, and the prediction is performed by:
\vspace{-3pt}
\begin{equation}
    \argmax_{i\in\left[C\right]} h_i(W_i,f(x)),
\label{eq:inference}
\end{equation}
\vspace{-3pt}
%
We consider four post processing methods in this paper including classifier normalization ($\tau$-norm~\cite{kang2019decoupling}), classifier re-scaling based on sample numbers (CDT-post~\cite{ye2020identifying, kim2020adjusting}), feature disentangling (TDE~\cite{tang2020longtailed}), and logit adjustment (LA-post~\cite{menon2020logit}), as shown in Table~\ref{tab:long_tail}.

%
\subsection{Analysis and Takeaways}
Here we summarize some of the key observations and knowledge revealed by the empirical study, including the effectiveness of LT methods on natural accuracy, the challenge to robustness evaluation reliability induced by some LT strategies, and the importance of boundary error for robust accuracy.

\noindent\textbf{Natural accuracy is easy to improve.}
According to the results in Table~\ref{tab:long_tail}, a number of LT strategies are proved effective in improving $A_{nat}$, as marked in green. We can either apply a cost-sensitive loss in outer minimization $\mathcal{L}_T$ during the whole training stage, or leverage fine-tuning and post-processing to boost the performance with little extra computational cost. 
This indicates that both re-balancing strategies applied during the training process and boundary adjustment at inference time positively impact $A_{nat}$, which inspires the development of our method in Sec.~\ref{sec:method}.

\noindent\textbf{Fake improvement exists for robust accuracy evaluation.}
It is noticeable that methods based on classifier normalization~\cite{kang2019decoupling} and re-scaling~\cite{ye2020identifying, kim2020adjusting} achieve impressive robust accuracy under PGD-attack, as marked in red, while the AA evaluated results remain ordinary. This is due to the sensitivity of PGD attack to both logits stretching and compressing, which is worth attention.

\begin{figure}[t]
	\centering
	\includegraphics[width=1.0\linewidth]{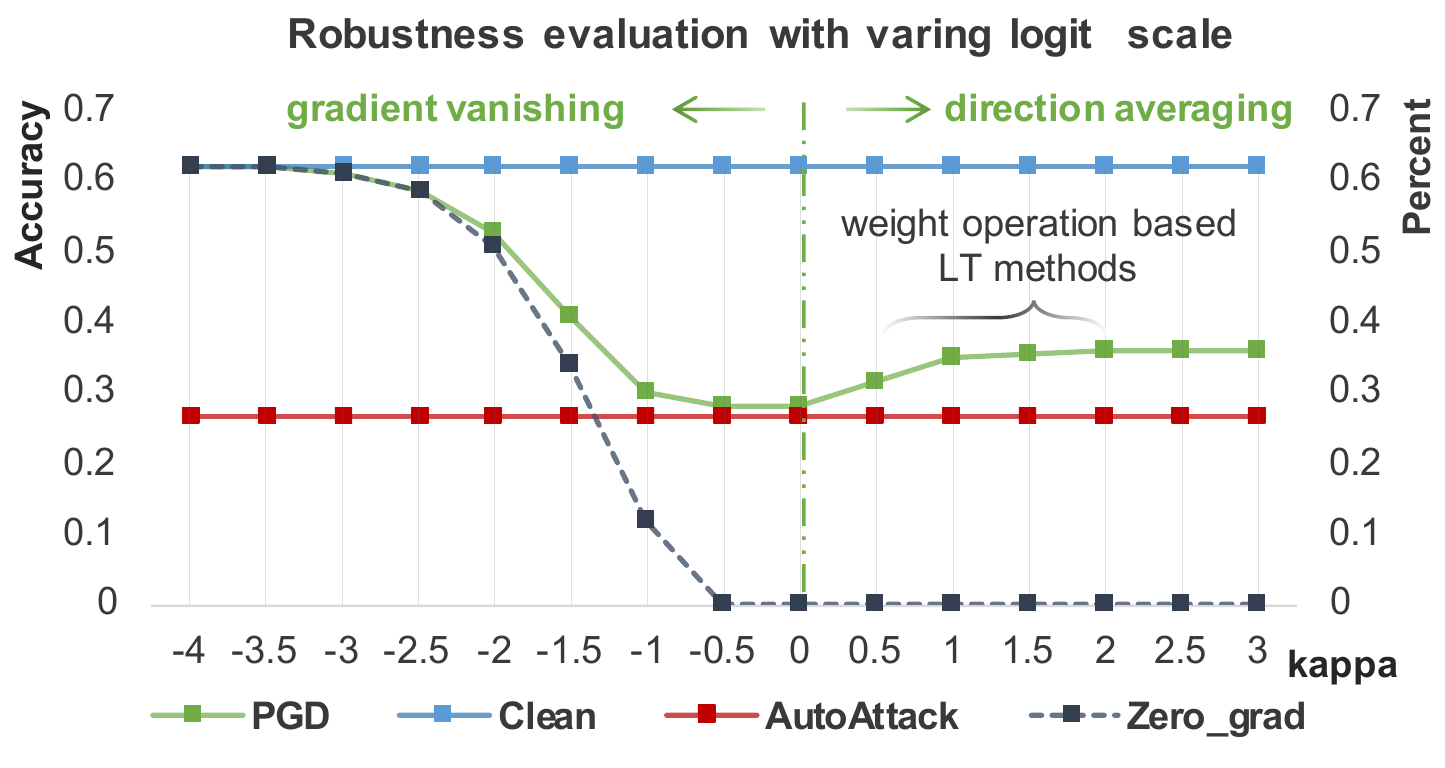}
	\caption{\small
	$A_{nat}$ and $A_{rob}$ under AA are invariant to logit scaling, while $A_{rob}$ under PGD is easily over estimated when using weight operation based LT methods.
	The zero in gradients occurs on originally correctly predicted images with logit stretching.
	}
	\label{fig:varying_logit_scale}
	\vspace{-12pt}
\end{figure}

Consider a uniformly re-scaled classifier $W_i'=W_i / 10^\kappa$ at inference time, where logit scales and $\kappa$ are negatively correlated. 
As shown in Fig.~\ref{fig:varying_logit_scale}, $A_{nat}$ is not effected since the re-scaling operation by $10^{-\kappa} > 0$ does not change the ordering; AA evaluated $A_{rob}$ is also invariant to scaling which can be seen as a reliable evaluation of robustness; while PGD robustness exhibits a minimum at $\kappa \approx 0$ and increases as $\kappa$ leaves zero on both sides.

The reason lies in the updating of PGD attack, which is based on the gradient produced by the inner CE loss: 
\begin{equation}
\begin{aligned}
    \nabla_{f(x)}\mathcal{L}_{CE}(W; f(x),y) & = -\nabla_{f(x)}z_y +\sum_{i}p_i\nabla_{f(x)}z_i \\
    & = \sum_{i\neq y}p_i(W_i - W_y).
\end{aligned}    
\label{eq:grad_ce}
\end{equation}
The vectors $W_y$ to $W_i$ with $i \neq y$ are weighted by \textit{softmax} produced prediction confidence, $p_i$, which is effective by focusing more on easily confused classes. 
But the sensitivity of softmax to both the absolute and relative values of its components leads to two kinds of fail cases of PGD. 

\textit{1) Gradient vanishing.}
The false sense of security due to gradient vanishing is not new knowledge~\cite{carlini2017CW, croce2020AA}:
a correctly predicted clean image with $y = \argmax_i{z_i}$ would gain $p_y \approx 1$ and $p_i \approx 0 (i \neq y)$ when all logits are scaled up, leading to zero gradient as in Eqn.~\ref{eq:grad_ce}.
We calculate the ratio of zero in gradient at pixel level during the PGD updating following~\cite{croce2020AA}, and it converges to the same level as $A_{nat}$ rather than 100\% in their paper (Fig.~\ref{fig:varying_logit_scale}).

\textit{2) Direction averaging.}
On the other side, compressed logits lead to averaged softmax outputs, where $\ p_i\approx 1/C$, and the updating direction becomes $\nabla_{f(x)}\mathcal{L}_{CE} = \frac{1}{C}\sum_iW_i-W_y = \overline{W}-W_y$, 
pointing from $W_y$ to the averaged weight vector $\overline{W}$. It only depends on $y$ and fails to take sample-specific properties into consideration; and since it is fixed throughout the inner maximization procedure of Eqn.~\ref{eq:AT}, the iterative attack actually degenerates to be single-step.
Consequently, the attack is weakened to some extent so that $A_{rob}$ gradually converges to a fixed value as $\kappa$ goes up, leading to fake gain of performance that weight operation based methods suffer from.

\noindent\textbf{Boundary error matters for robust accuracy.}
When erasing the false sense of security, the reliable $A_{rob}$ under AA in Table~\ref{tab:long_tail} seems not significantly affected as $A_{nat}$ raises. In fact, a huge gap between natural and robust accuracy is observed, and it continuously widens as the former improves, which can be reflected by the term $R_{bdy}$\footnote{The case that a wrong prediction being “corrected" after the attack rarely happens, so $\mathbf{R}_{bdy}$ can be basically reflected by $A_{nat} - A_{rob}$.} in Eqn.~\ref{eq:error_decomposition}.
The phenomenon exposes the importance of controlling $R_{bdy}$ in order to improve both $A_{nat}$ and $A_{rob}$, which is an issue that many LT methods do not promise to solve.
However, we found that cosine classifier based methods could benefit from a relatively smaller $R_{bdy}$ compared with linear classifiers. One evidence is that in Table~\ref{tab:long_tail}, the models trained with a cosine classifiers exhibit the lowest gap between $A_{nat}$ and $A_{rob}$ under AA, marked in blue in the last column.
We would analyze the reason behind it in the next section and how to take advantage of its good property.
\section{Methodology}
\label{sec:method}

Earlier discoveries cast light on two key factors of solving this challenging problem:
1) a proper feature and classifier embedding helps to achieve a lower boundary error $R_{bdy}$,
and 2) the combination of long-tailed recognition (LT) methods with adversarial training (AT) framework would benefit $A_{nat}$. 
Hence, we propose a clean yet effective approach which consists of two components, \ie scale-invariant classifier and two-stage re-balancing, to achieve \textbf{Ro}bust and \textbf{Bal}anced predictions, namely \textbf{RoBal}.


\subsection{Scale-invariant Classifier} 

 \begin{figure}[t]
	\centering
	\includegraphics[width=0.98\linewidth]{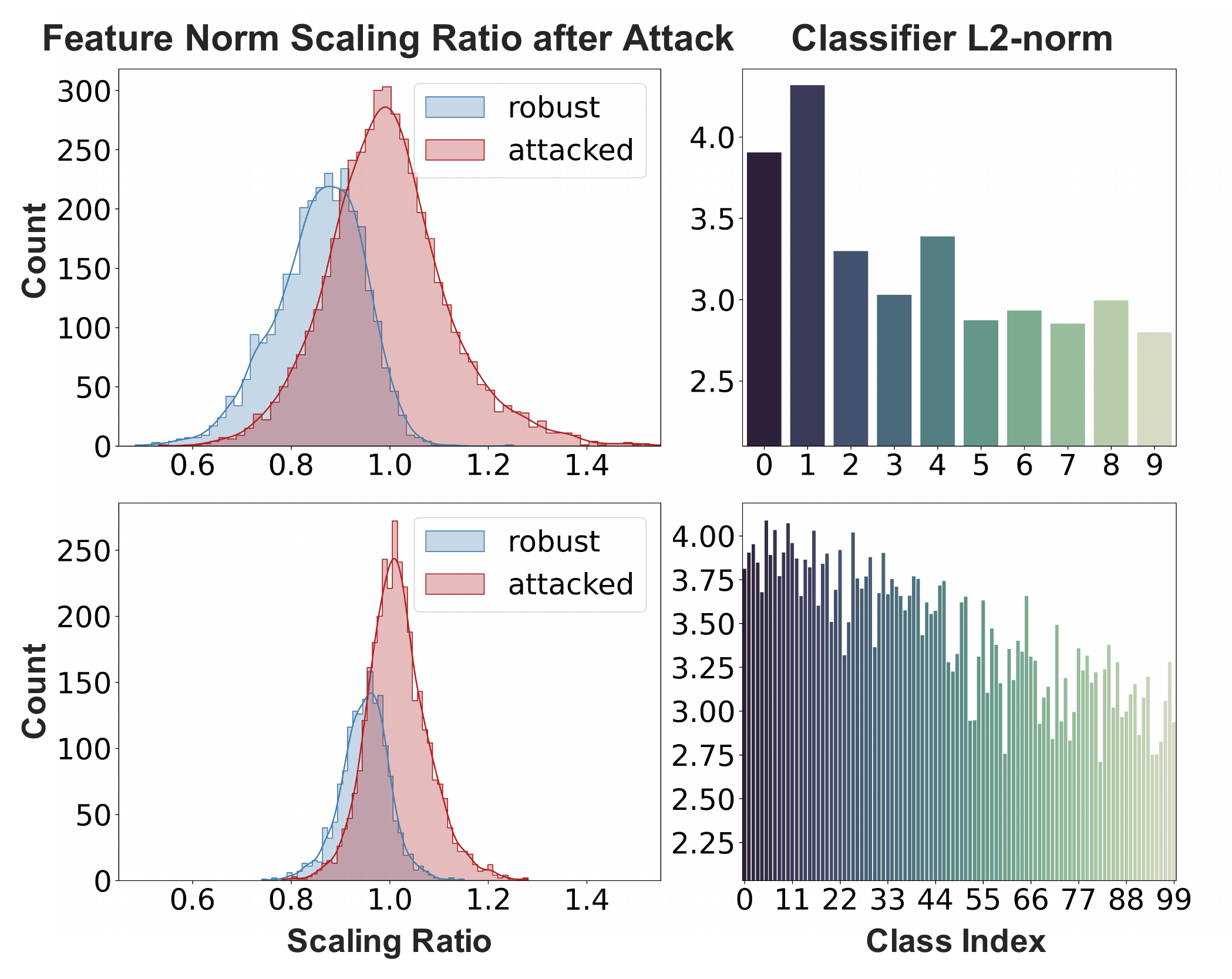}
	\caption{\small
	\textbf{Left:} the feature norm distribution shifts after adversarial attack, the successfully attacked samples statistically have a larger scaling factor than those that stay robust;
	\textbf{Right:} the classifier weight norm roughly decreases towards the tail classes.
	}
	\vspace{-10pt}
	\label{fig:pgd_scaling}
\end{figure}

In a basic classification task using a standard linear classifier, the predicted logit of class $i$ can be represented as:
\begin{equation}
    W_i^T f(x) + b_i = || W_i ||\cdot || f(x) || \cos{\theta_i} + b_i,
    \label{eq:fc}
\end{equation}
where we can see that the prediction depends on three factors: 1) the scale of weight vector $|| W_i ||$ and feature vector $|| f(x) ||$; 
2) the angle $\cos{\theta_i}$ between them; 
and 3) the bias of the classifier $b_i$.
In this section we would focus on the first factor to show the importance of being scale-invariant.

Firstly, the decomposition above indicates that the prediction of a sample can be changed by simply scaling its norm in the feature space.
We consider this to be one of the schemes adversarial examples use to confuse the model,
leading to different feature norm distributions between ``successfully attacked'' and ``robust'' samples.
To be specific, the originally correctly predicted images can be separated into two groups by whether the attack is successful. We then calculate the scaling ratio $\left\Vert f(x+\delta) \right\Vert / \left\Vert f(x) \right\Vert $ between each attacked and clean input pair. We could observe different distributions of the two groups in Fig.~\ref{fig:pgd_scaling}, where a successful attack is more likely to happen with a relatively higher scaling ratio.

\begin{figure}[t]
	\centering
	\includegraphics[width=1.0\linewidth]{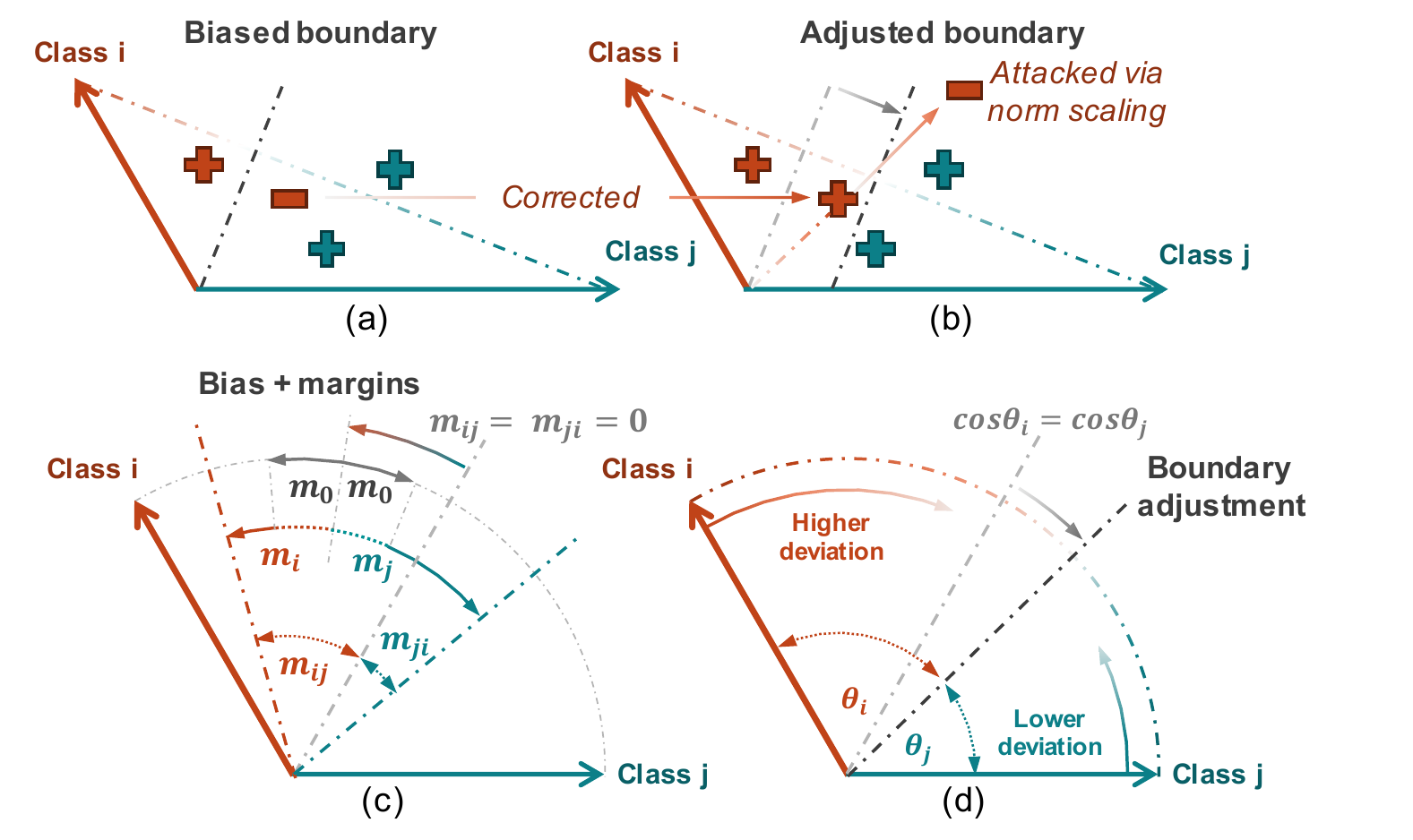}
	\caption{\small
	\textbf{(a)} and \textbf{(b)}: biased decision boundary is induced by imbalanced weight norms while boundary adjustment helps correct some mistakes; feature norm scaling can result in a successful attack;
	\textbf{(c)}: margin construction with \textit{hyper-parameters ignored} for simplicity;
	\textbf{(d)}: boundary adjustment at inference stage.
	}
	\vspace{-10pt}
	\label{fig:fc_and_margin}
\end{figure}

Besides the feature embedding, the scales of the weight vectors $||W_i||$ in a linear classifier would also induce problems to the long-tail scenario. Specifically, they usually decrease towards the tail classes as shown in Fig.~\ref{fig:pgd_scaling}, which is also observed in previous works~\cite{kang2019decoupling, kim2020adjusting}. 
Different weight scales result in biased decision boundaries (Fig.~\ref{fig:fc_and_margin}) and hurt recognition performance. This could be alleviated by adjusting the class-specific bias. But it still suffers from a high adversarial risk even after boundary adjustment considering the varying feature norm, as shown in Fig.~\ref{fig:fc_and_margin}

Based on the observation and analysis, a natural idea is to remove the influence of scales from both the features and weights. Therefore, a scale-invariant classifier, \eg, cosine classifier that limits the vectors to a hyper-sphere~\cite{pang2020boosting}, could be a proper choice.
The benefit of it reducing $R_{bdy}$ has been revealed in Table~\ref{tab:long_tail}.
And we would further introduce how to address the issue of imbalance via its combination with re-balancing strategies.

\subsection{Two-stage Re-balancing}
%
Based on the normalized features and classifier weights, we further consider the problem of long-tailed data distribution.
Recall the knowledge we gain from Sec.~\ref{sec:empirical_study}, where significant natural accuracy gain via effective elimination of imbalance can happen either during training or simply at inference time.
Accordingly, we propose a two-stage re-balancing framework in this section, which exactly focuses on the other two factors in Eqn.~\ref{eq:fc}, namely $\cos{\theta_i}$ and $b_i$:
1) margins introduced to the cosine classifier at training stage promote a more compact representation learning, where both class-aware and pair-aware margin engineering would boost network performance in our long-tailed scenario;
2) boundary adjustment at inference stage further tackles the issue of higher variance and deviation in tail classes.
Different cases of the compensation and cooperation between them is explored in ablation study in Sec.~\ref{sec:experiment}.

\begin{table*}[htbp]
  \centering
  \caption{Experimental results on CIFAR-10-LT and CIFAR-100-LT with WideResnet-34-10.}
  \small
    \begin{tabular}{p{5.2em}|c|ccccc||c|ccccc}
    \toprule
    \multicolumn{1}{c|}{\textbf{Dataset}} & \multicolumn{6}{c||}{\textbf{CIFAR-10-LT}}    & \multicolumn{6}{c}{\textbf{CIFAR-100-LT}} \\
    \midrule
    \textbf{Methods} & \multicolumn{1}{p{2em}|}{\textbf{Clean}} & \multicolumn{1}{p{2em}}{\textbf{FGSM}} & \multicolumn{1}{p{2em}}{\textbf{PGD}} & \multicolumn{1}{p{2em}}{\textbf{MIM}} & \multicolumn{1}{p{2em}}{\textbf{CW}} & \multicolumn{1}{p{2em}||}{\textbf{AA}} & \multicolumn{1}{p{2em}|}{\textbf{Clean}} & \multicolumn{1}{p{2em}}{\textbf{FGSM}} & \multicolumn{1}{p{2em}}{\textbf{PGD}} & \multicolumn{1}{p{2em}}{\textbf{MIM}} & \multicolumn{1}{p{2em}}{\textbf{CW}} & \multicolumn{1}{p{2em}}{\textbf{AA}} \\
    \midrule
    Plain & 77.16 & 7.01  & 0.00  & 0.00  & 0.00  & 0.00  & 62.29 & 3.94  & 0.00  & 0.00  & 0.00  & 0.00 \\
    \midrule
    \multicolumn{1}{l|}{AT~\cite{madry2018towards}} 
    & 62.33 & 33.57 & 29.30 & 30.02 & 30.31 & 28.15 
    & 48.96 & 21.06 & 17.26 & 17.80 & 17.65 & 16.26 \\
    \multicolumn{1}{l|}{TRADES~\cite{zhang2019trades}} 
    & 54.29 & 32.80 & 30.20 & 30.53 & 29.58 & 28.94 
    & 43.71 & 23.06 & \textbf{21.13} & \textbf{21.42} & 19.49 & 18.68 \\
    \multicolumn{1}{l|}{HE~\cite{pang2020boosting}} 
    & 58.47 & 35.17 & 31.73 & 32.26 & 29.96 & 28.45 
    & 48.63 & 23.06 & 19.56 & 20.28 & 19.20 & 17.60 \\
    \multicolumn{1}{l|}{MMA~\cite{ding2020mma}} 
    & 61.51 & 36.40 & 29.29 & 30.38 & 29.59 &  25.91 
    & \textbf{54.98}   & 19.65 & 13.52 & 13.98 & 12.35 & 14.54 \\
    \multicolumn{1}{l|}{AVmixup~\cite{lee2020adversarial}} 
    & 66.97 & 33.90 & 28.40 & 29.67 & 26.43 & 24.39 
    & 52.45 & 23.28 & 19.04 & 20.78 & 14.82 & 12.60 \\
    \midrule
    \textit{RoBal-N}
    & \textbf{75.52} & 40.04 & 33.50 & 34.57 & 33.68 & 31.72  
    & 51.63 & 22.81  & 19.01 & 19.50 & 19.42 & 18.16 \\
    \textit{RoBal-R} 
    & 74.51 & \textbf{40.55} & \textbf{33.87} & \textbf{34.92} & \textbf{34.12} & \textbf{32.04}  
    & 50.38 & \textbf{23.59} & 19.48 & 20.13 & \textbf{20.16} & \textbf{18.69} \\

    \bottomrule
    \end{tabular}%
  \label{tab:defense}%
  \vspace{-10pt}
\end{table*}%

\noindent\textbf{Class-aware Margin on Training.} 
A straight forward and widely used idea to take class imbalance into consideration is to assign class-specific bias in the CE loss during training. Following Ren~\etal~\cite{ren2020balanced-softmax} and Menon~\etal~\cite{menon2020logit}, we adopt the form of $b_i = \tau_b\log(n_i)$, and the modified CE Loss becomes:
\begin{equation}
\begin{aligned}
   \mathcal{L}_0 & = -\log(\frac{e^{z_y+b_y}}{\sum_i e^{z_i+b_i}})
   = \log(1 + \sum_{i\neq y}e^{z_i - z_y + \tau_b\log(\frac{n_i}{n_y})}),
\end{aligned}
\label{eq:loss_ce_bias}
\end{equation}
where $\tau_b$ is a hyper-parameter controlling the bias value calculation.
However, on considering the formulation in the manner of margin, we find that the margin from the ground truth class $y$ to class $i$, namely $\tau_b\log(n_i/n_y)$, would become negative when $n_y > n_i$, leading to less discriminating representation and classifier learning of head classes. 
To deal with the issue, we further add a class-aware margin term together with the pre-defined $b_i$, which assigns a larger margin value to the head class in compensation:
\begin{equation}
   m_i = \frac{\tau_m}{s}\log{\frac{n_i}{n_{min}}} + m_0.
\label{eq:b_i}
\vspace{-4pt}
\end{equation}
Here the first term would increase along with with $n_i$ while achieving its lowest at zero when $n_i = n_{min}$, and $\tau_m$ is the hyper-parameter to control the trend;
the second term $m_0 > 0$ is a uniform margin for all classes, as is a commonly used strategy for cosine classifier based networks~\cite{wang2018cosface};
$s$ represents a temperature here to expand the value range of the cosine outputs, which helps to present a more clearly formulated loss function as below:
\begin{equation}
\begin{aligned}
    \mathcal{L}_1 & = -\log\left(\frac{e^{s (\cos{\theta_y} - m_y) + b_y}}{e^{s (\cos{\theta_y} - m_y) + b_y} + \sum_{i \neq y} e^{s\cos{\theta_i} + b_i}}\right) \\
    & = \log(1 + \sum_{i\neq y}e^{s(\cos{\theta_i} - \cos{\theta_y} + m_{yi})}),
\end{aligned}
\label{eq:loss_cosine_ce_1}
\end{equation}
where $\cos{\theta_i} = \widetilde W_i^T \widetilde f(x)$, and
\begin{equation}
\begin{aligned}
    m_{yi} & = \frac{\tau_b}{s}\log(\frac{n_i}{n_y}) + \frac{\tau_m}{s}\log(\frac{n_y}{n_{min}}) + m_0 \\
    & = \frac{(\tau_b-\tau_m)}{s}\log(\frac{n_i}{n_y}) +  \frac{\tau_m}{s}\log(\frac{n_i}{n_{min}}) + m_0.
\end{aligned}
\label{eq:loss_cosine_ce_2}
\end{equation}
Notice that we here adopt $\widetilde W_i = W_i / (||W_i|| + \gamma)$ here, which is slightly different from Sec.~\ref{sec:empirical_study} while we empirically find it able to produce slightly better performance.
The \textbf{first} line of formulation in Eqn.~\ref{eq:loss_cosine_ce_2} constructs the margin between ground truth class $y$ and a negative class $i$: a composition of a pair-aware margin $\log(n_i/n_y)$ scaled by $\tau_b$, a class-aware margin $\log(n_i/n_{min})$ scaled by $\tau_m$, and a uniform $m_0$, as shown in Fig.~\ref{fig:fc_and_margin}.
While The \textbf{second} line reveals a more direct relationship to the data distribution: $n_y$ occurs only in the first term to assign a larger margin to tail classes when $\tau_b - \tau_m > 0$. It encourages a more compact and discriminating learning on them and especially benefits the imbalance learning process. We would show how each term effects the training in the ablation study.

\noindent\textbf{Class-specific Bias on Inference.}
With a normalized classifier, the decision boundary is naturally unbiased at inference time.
However, the sparse data distribution in the tail leads to higher uncertainty~\cite{khan2019striking} and feature deviation~\cite{ye2020identifying}, while head classes would benefit from a more compact feature embedding and concise classifier learning.
As a result, when using a uniform margin $m_0$ with $\tau_b = \tau_m = 0$ during training, we can still observe an obvious decrease in the recall of the tail classes. 
Thus a post processing strategy to adjust the cosine boundary is still needed, as can be formulated as a dual process to the pre-defined $b_i$ during training in Eqn.~\ref{eq:b_i}. $\tau_p$ is introduced here and the inference becomes:
\begin{equation}
    \argmax_{i\in [C]}{ s\cdot\cos{\theta_i} - \tau_p \log(\frac{n_i}{\sum_j{n_j}})}
    \vspace{-3pt}
\end{equation}
Actually, when class-specific margins are added along with the uniform one, the dependence on boundary adjustment is eliminated, as to be explored in Sec.~\ref{sec:experiment}.

\noindent\textbf{Regularization Term.}
Finally, inspired by the some of the well-known AT variants~\cite{kannan2018alp, zhang2019trades, ding2020mma}, an additional regularization term between the paired features or logits produced by the clean and perturbed images would further promote the robustness performance. Different kinds of regularization terms can be easily adopted into the training framework via modification on the loss function, and we follow Zhang~\etal~\cite{zhang2019trades} to take advantage of a KL-divergence term, and the overall loss function become:
\vspace{-2pt}
\begin{equation}
    \mathcal{L} = \mathcal{L}_1(x+\delta,y) + \alpha \cdot KL(~\widetilde W_i^T \widetilde f(x+\delta), ~\widetilde W_i^T \widetilde f(x))
    \label{eq:final}
\end{equation}
where $\delta$ is the perturbation generated by inner maximization guided by a plain cross-entropy loss, performed on the direct outputs of the cosine operation without margins.
\vspace{-3pt}


\section{Experiments}
\label{sec:experiment}
\vspace{-3pt}

\noindent\textbf{Datasets.}
We conduct experiments on the long-tailed versions of CIFAR-10 and CIFAR-100 following~\cite{cui2019cb}.
Imbalance Ratio (IR) in the main experiments is set as 50 and 10, 
respectively.
Experimental results with various IRs are also provided in Table~\ref{tab:imbalance_ratio}.

\noindent\textbf{Evaluation Metrics.}
On evaluating model robustness, the allowed $l_\infty$ norm-bounded perturbation is $\epsilon=8/255$. Attacks conducted include the single-step attack FGSM~\cite{goodfellow2015FGSM} and several iterative attacks including PGD, MIM, and C\&W performed for 20 steps with a step size of $2/255$.
We also use the recently proposed Auto Attack (AA)~\cite{croce2020AA} which is an ensemble of different attacks and is parameter-free.

\noindent\textbf{Comparison Methods.}
We compare our method with several state-of-the-art defense methods besides the standard AT~\cite{madry2018towards}, including
TRADES~\cite{zhang2019trades}, MMA~\cite{ding2020mma}, HE~\cite{pang2020boosting}, and AVmixup~\cite{lee2020adversarial},
among which AVmixup~\cite{lee2020adversarial} is re-implemented and the others are evaluated with the officially released code.
\textbf{Implementation details} on our network training, hyper-parameter setting, and attacking algorithms are included in the supplementary material.

\subsection{Comparison Results}
The comparison with other defense methods is reported in Table~\ref{tab:defense}. Since a trade-off between \textbf{N}atural and \textbf{R}obust accuracy usually exists, we report our results with different emphasis, denoted by \textit{RoBal-N} and \textit{RoBal-R}, respectively. The setting of hyper-parameters to control the trade-off can be found in Sec.~\ref{supp:sec:experiments}.
On CIFAR-10-LT, our method significantly outperforms all the compared ones on both $A_{nat}$ of clean images and $A_{rob}$ under five different attacks.
On CIFAR-100-LT, TRADES~\cite{zhang2019trades} and RoBal achieve comparable results on robust accuracy.
However, they significantly sacrifice the performance on $A_{nat}$, while our method also consistently boosts $A_{nat}$ compared with AT baseline. AVmixup~\cite{lee2020adversarial} and MMA~\cite{ding2020mma} (on CIFAR-100-LT) achieve decent $A_{nat}$ while suffering from the poor robust performance under AA.
It is noted that the overall improvement in CIFAR-100-LT is less significant than CIFAR-10-LT, possibly due to the smaller imbalance ratio or the increase of class number; thus we also provide preliminary results on ImageNet-LT~\cite{liu2019largescale} with 1000 classes in Sec.~\ref{supp:sec:imagenet-lt}.

%

\subsection{Ablation Study}
%
Here we explore the effect of hyper-parameters during the two-stage re-balancing.
We mainly discuss three terms according to Eqn.~\ref{eq:loss_cosine_ce_2}, namely  $m_0$, $\tau_b-\tau_m$, and $\tau_m$. We change the critical variable while fix the others for analysis as shown in Table~\ref{tab:margin}. Several interesting observations include: 
1) a proper $m_0 > 0$ leads to higher $A_{rob}$ yet lower $A_{nat}$ at both stages;
2) a higher $\tau_b-\tau_m$ promotes natural accuracy at the end of training stage, yet a peak is observed for robustness;
3) a larger $\tau_m$ helps reduce the boundary error $R_{bdy}$, while it could hurt $A_{nat}$ if set too large.
\textit{Please refer to the supplementary for more experimental results.}
\begin{table}[htbp]
  \centering
      \caption{Effect of hyper-parameters. We fix $\tau_m=\tau_b=0$ in block 1, $m_0=0.1, \tau_m=0$ in block 2, and $m_0=0.1,\tau_b-\tau_m=1.2$ in block 3. ** denotes the key metric that is worth noting.}
     \small
    \begin{tabular}{c|cc||cc}
    \toprule
    \multicolumn{3}{c||}{\textbf{End of training stage}} & \multicolumn{2}{c}{\textbf{Inference with $\tau_p*$}} \\
    \midrule
    \textbf{$m_0$} & \textbf{Clean} & \textbf{AA} & \textbf{Clean**} & \textbf{AA} \\
    \midrule
    0     & 63.51 & 28.47 & 75.98 & 30.46 \\
    0.1   & 63.17 & 29.13 & 75.51 & 31.24 \\
    0.2   & 60.6  & 29.04 & 74.83 & 32.04 \\
    0.3   & 56.59 & 28.63 & 71.31 & 31.28 \\
    \midrule
    \textbf{$\tau_b - \tau_m$} & \textbf{Clean**} & \textbf{AA} & \textbf{Clean} & \textbf{AA} \\
    \midrule
    0     & 63.51 & 28.47 & 75.51 & 31.24 \\
    0.5   & 68.58 & 30.55 & 75.77 & 30.94 \\
    1     & 73.93 & 31.52 & 75.42 & 31.61 \\
    1.5   & 74.67 & 30.95 & 74.67 & 30.95 \\
    \midrule
    \textbf{$\tau_m$} & \textbf{Clean} & \textbf{AA} & \multicolumn{2}{c}{\textbf{Training  gap**}} \\
    \midrule
    -0.3  & 74.88 & 30.66 & \multicolumn{2}{c}{44.22} \\
    0     & 75.08 & 31.79 & \multicolumn{2}{c}{43.29} \\
    0.3   & 74.51 & 32.04 & \multicolumn{2}{c}{42.47} \\
    0.6   & 71.84 & 31.41 & \multicolumn{2}{c}{40.43} \\
    \bottomrule
    \end{tabular}%
  \label{tab:margin}%
\end{table}%


\begin{table}[t]
  \centering
  \caption{Experimental results on CIFAR-10-LT with different IRs}
  \small
    \begin{tabular}{c|c|c|cccc}
    \toprule
    \textbf{IR}    & \textbf{Methods} & \textbf{Clean} & \textbf{PGD}   & \textbf{MIM}  & \textbf{CW}    & \textbf{AA} \\
    \midrule
    \multirow{3}[2]{*}{100} & AT    & 56.72 & 27.27 & 27.87 & 27.80 & 25.97 \\
          & TRADES & 45.67 & 26.97 & 27.29 & 26.53 & 25.93 \\
          & Our   & \textbf{68.07} & \textbf{30.35} & \textbf{31.25} & \textbf{30.55} & \textbf{28.97} \\
    \midrule
    \multirow{3}[2]{*}{50} & AT    & 62.33 & 29.30 & 29.72 & 30.31 & 28.15 \\
          & TRADES & 54.29 & 30.20 & 30.53 & 29.58 & 28.94 \\
          & Our   & \textbf{73.93} & \textbf{34.24} & \textbf{35.37} & \textbf{34.58} & \textbf{32.70} \\
    \midrule
    \multirow{3}[2]{*}{20} & AT    & 74.09 & 33.59 & 34.=65 & 34.27 & 32.02 \\
          & TRADES & 65.17 & 34.65 & 35.29 & 34.07 & 33.06 \\
          & Our   & \textbf{78.49} & \textbf{39.17} & \textbf{40.44} & \textbf{39.38} & \textbf{37.58} \\
    \midrule
    \multirow{3}[2]{*}{10} & AT    & 79.45 & 37.11 & 37.93 & 38.31 & 35.51 \\
          & TRADES & 72.92 & 39.15 & 39.95 & 38.41 & 37.33 \\
          & Our   & \textbf{81.20} & \textbf{40.22} & \textbf{41.75} & \textbf{40.91} & \textbf{38.90} \\
    \bottomrule
    \end{tabular}%
  \label{tab:imbalance_ratio}%
  \vspace{-5pt}
\end{table}%

\vspace{-15pt}

\begin{figure}[t]
	\centering
	\includegraphics[width=0.90\linewidth]{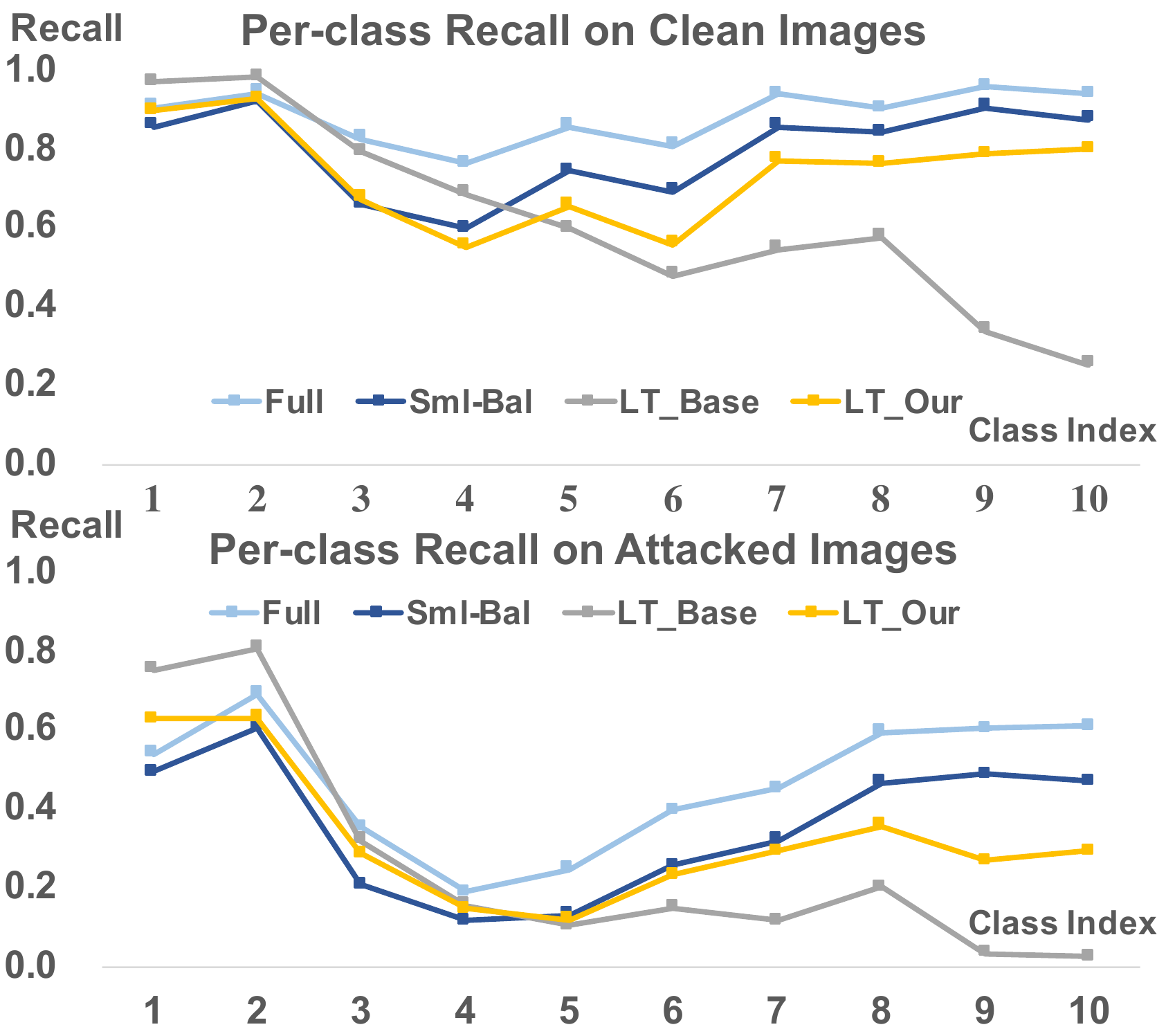}
	\caption{\small
	Results under different data scales and distributions. 
	}
	\vspace{-10pt}
	\label{fig:recall_precision}
\end{figure}

\subsection{Further Analysis}

\noindent\textbf{Effect of Data Scale.}
Since a long-tailed dataset differs from the full dataset by both class-wise sample numbers and the overall data scale, we conduct a comparison with 1) the original full dataset (Full) and 2) a dataset with the same number of samples as the long-tailed version but is uniformly distributed (Sml-Bal). 
From the per-class recall shown in Fig.~\ref{fig:recall_precision},
we can see that:
(1) Performance of Sml-Bal are uniformly lower than that of the full dataset.
(2) The basic AT framework produces an apparent decrease in recall from head to tail on both clean and attacked images. Specifically, head classes gain even higher robustness than balanced baselines, indicating that the intrinsic prediction bias raise their resistance to attack. 
(3) Our RoBal (LT-Our) applied to the long-tailed dataset efficiently re-balances the per-class recall compared with the baseline (LT-base).
%

\noindent\textbf{Effect of Imbalance Ratio.}
We also constructed long-tailed datasets with different imbalanced ratios (\textbf{IR}) following~\cite{cui2019cb} to evaluate the performance of AT, TRADES~\cite{zhang2019trades}, and our methods.
As shown in Table~\ref{tab:imbalance_ratio}, our method outperforms the baseline and TRADES~\cite{zhang2019trades} remarkably on both natural accuracy and robust accuracies over different IRs.
%

\vspace{-5pt}
\section{Conclusion}
\label{sec:conclusion}

In this paper, 1) we first reveal the negative impacts induced by long-tailed data distribution on both recognition performance and adversarial robustness, uncovering the intrinsic challenges of this problem.
2) Then, a systematic study on existing long-tailed recognition approaches and their combination with the adversarial training framework contributes several valuable observations. 
3) Finally, inspired by them, we propose a clean yet effective framework that benefits from the norm-invariant property of cosine classifier and a two-stage re-balancing framework, which outperforms existing state-of-the-art defense methods.
To our best knowledge, we are the first to tackle adversarial robustness under long-tailed distribution, which we believe would be a significant step towards real-world robustness.

\vspace{-5pt}
\paragraph{Acknowledgements.}
This research was conducted in collaboration with SenseTime. This work is supported by GRF 14203518, ITS/431/18FX, CUHK Agreement TS1712093, NTU NAP and A*STAR through the Industry Alignment Fund - Industry Collaboration Projects Grant, and the Shanghai Committee of Science and Technology, China (Grant No. 20DZ1100800).

\appendix
\renewcommand{\thetable}{S\arabic{table}}
\renewcommand\thefigure{S\arabic{figure}}


\section{Implementation Details of Experiments}
\label{supp:sec:experiments}
\subsection{Training Details and Hyper-parameter Setting}
We adopt the WideResNet-34-10 as the model architecture. The initial learning rate is set as 0.1 with a decay factor of 10 at 60 and 75 epochs, totally 80 epochs. We use the last epoch for evaluation without early-stop for all the methods. We use the SGD momentum optimizer with weight decay set as $2\times10^{-4}$. 
We use a batch size of 64 for all the experiments in the main paper.
The adversarial training is applied with the maximal permutation of $8/255$ and a step size of $2/255$ ($0.031$ and $0.0078$ are used for implementation). The number of iterations in the inner maximization is set as $5$, and a study on the effect of PGD steps in AT is reported in Sec.~\ref{supp:sec:pgd_step}.
There are multiple hyper-parameters involved, where those that control margins or boundary adjustment are the most critical.
Specifically, we adopt $m_0 = 0.1$ for CIFAR-10-LT and $m_0 \in \{0.2,0.3\}$ for CIFAR-100-LT for different emphasis (i.e., the trade-off between natural and robust accuracy). 
$\tau_b-\tau_m=1.2$ in Eqn.10 would basically produce a good result via training stage re-balancing, while $\tau_b-\tau_m=0$ with $\tau_p=1.5$ would also work well based on pure boundary adjustment at inference time. The optimal value of $\tau_p$ relies mainly on $\tau_b - \tau_m$. The ablation study includes detailed comparisons.
Other hyper-parameters are less sensitive and have relatively small impact on the performance, where we adopt $s = 10$, $\gamma \in \{1/32, 1/16\}$, and we set $\alpha=6,3$ in Eqn.12 for CIFAR-10-LT and CIFAR-100-LT, respectively. 


%
\subsection{Code References}
For the defense methods we compare with, we leverage the officially released code for them if available, including
TRADES~\cite{zhang2019trades}~\footnote{\url{https://github.com/yaodongyu/TRADES}},
MMA~\cite{ding2020mma}~\footnote{\url{https://github.com/BorealisAI/mma_training}},
Free~\cite{shafahi2019free}~\footnote{\url{https://github.com/mahyarnajibi/FreeAdversarialTraining}},
and HE~\cite{pang2020boosting}~\footnote{\url{https://github.com/ShawnXYang/AT_HE}}.
AVmixup~\cite{lee2020adversarial} are re-implement according to the paper.

For the attacks used for evaluation, we refer to several officially released code bases and the original papers for the implementation, including FGSM~\cite{goodfellow2015FGSM}, PGD~\cite{madry2018towards}, MIM~\cite{dong2018MIM}, C\&W~\cite{carlini2017CW}, and Auto Attack~\cite{croce2020AA}~\footnote{\url{https://github.com/fra31/auto-attack}}.

For the long-tailed recognition methods in Table 1, we also refer to the official code of them if available.

\subsection{Implementation Details of Table 1}
\label{supp:subsec:implementation_detail_LT}
In Sec.3.2 of the paper, we revisit and formulate a number of long-tailed recognition methods. We would report the hyper-parameters selected for them when combining with adversarial training framework in our implementation in Table 1, where we choose the \textbf{optimal values} by searching the hyper-parameters with a step size of $1$ or $0.1$.
%
\begin{table*}[htbp]
  \centering
  \caption{Hyper-parameters selected for LT methods used in Table 1, where we choose the \textbf{optimal values} by searching the hyper-parameters with a step of $1$ or $0.1$. * denotes that we use CB-Focal.}
  \small
    \begin{tabular}{c|l|l|l}
    \toprule
    \toprule
    \multicolumn{1}{c|}{\textbf{Stage}} & \multicolumn{1}{c|}{\textbf{Methods}} & \multicolumn{1}{c|}{\textbf{Formulation}} & \multicolumn{1}{c}{\textbf{Hyper-parameters}} \\
    \midrule
    \multirow{9}[18]{*}{Training} & Vanilla FC & $g_i = W_i^Tf(x)$ & - \\
\cmidrule{2-4}          & Vanilla Cos & $g_i = \widetilde W_i^T \widetilde f(x)$ & \textit{temperature} $s=16$ \\
\cmidrule{2-4}          & Class-aware margin~\cite{cao2019ldam} & $g_i = W_i^Tf(x) - \mathbbm{1}\{i=y\} \cdot \delta_i$ & $\delta_{max}=0.5, \delta \propto n^{-1/4}$ \\
\cmidrule{2-4}          & Cosine with margin~\cite{wang2018cosface, pang2020boosting} & $g_i = \widetilde W_i^T \widetilde f(x) - \mathbbm{1}\{i=y\} \cdot m$ & $m=0.2, s=10$ \\
\cmidrule{2-4}          & Class-aware temperature~\cite{ye2020identifying} & $g_i = W_i^Tf(x)\cdot (n_i / n_{max})^\gamma$ & $\gamma=0.3$ \\
\cmidrule{2-4}          & Class-aware bias~\cite{menon2020logit, ren2020balanced-softmax} & $g_i = {W_i}^Tf(x) + \tau \log(n_i)$ & $\tau=1$ \\
\cmidrule{2-4}          & Hard-example mining~\cite{lin2017focal} & $r(y) = \left( 1-p_y \right)^\gamma$, applyed with BCE loss & $\gamma=2$ \\
\cmidrule{2-4}          & Re-sampling~\cite{shen2016relay} & $r_s(i) \propto 1/n_i$    & - \\
\cmidrule{2-4}          & Re-weighting*~\cite{cui2019cb} & $r(y) = (1-\beta) / (1 - \beta^n_y)$ & $\beta=0.9999, \gamma=2$ \\
    \midrule
    \multicolumn{1}{c|}{\multirow{3}[6]{*}{Fine-tuning}} & One-epoch re-sampling~\cite{kang2019decoupling} & $h_i = {W'}_i^Tf(x)$, ${W'}_i$ re-trained with RS & - \\
\cmidrule{2-4}          & One-epoch re-weighting~\cite{cao2019ldam, cui2019cb} & $h_i = {W'}_i^Tf(x)$, ${W'}_i$ fine-tuned with RW & $\beta=0.9999, \gamma=2$ \\
\cmidrule{2-4}          & Learnable classifier scale~\cite{kang2019decoupling} & $h_i = s_i \cdot W_i^Tf(x)$, where $s_i$ is learnable & - \\
    \midrule
    \multicolumn{1}{c|}{\multirow{4}[8]{*}{Inference}} & Classifier re-scaling~\cite{ye2020identifying, kim2020adjusting} & $h_i = (W_i / n_i^\tau)^Tf(x)$ & $\tau=0.3$ \\
\cmidrule{2-4}          & Classifier normalization~\cite{kang2019decoupling} & $h_i = (W_i / \left\Vert W_i \right\Vert^\tau)^Tf(x)$ & $\tau=2$ \\
\cmidrule{2-4}          & Class-aware bias~\cite{menon2020logit} & $h_i = W_i^Tf(x) - \tau \log(n_i)$ & $\tau=1$ \\
\cmidrule{2-4}          & Feature disentangling~\cite{tang2020longtailed} & $h_i = W_i^T (f(x)-\alpha \cos(f(x),d)\cdot d)$   & $\alpha=0.1$ \\
    \bottomrule
    \bottomrule
    \end{tabular}%
  \label{tab:long_tail_param}%
\end{table*}%

\section{Extensive Experiments}
\label{supp:sec:extensive}
\subsection{Loss Functions in Adversarial Training}
In Sec.3, a modified loss function $\mathcal{L'}_{CE}$ can be adopted to AT procedure in three modes: replacing the CE in $\mathcal{L}_A$, $\mathcal{L}_T$, or both of them. We study the effect of the three modes in Table~\ref{tab:loss_function}.
It can be observed that:
1) replacing CE in $\mathcal{L}_A$ of the inner maximization would slightly benefit the natural accuracy with re-weighting~\cite{cui2019cb}, class-aware temperature~\cite{ye2020identifying}, and bias~\cite{menon2020logit, ren2020balanced-softmax}, while re-weighting would hurt robustness in this scenario; class-aware margin~\cite{cao2019ldam} is beneficial to robust accuracy but hurts the natural accuracy slightly;
2) replacing  CE in $\mathcal{L}_T$ of the outer minimization or both $\mathcal{L}_A$ and $\mathcal{L}_T$ would result in a significantly higher natural accuracy with class-aware temperature and bias, and the robust accuracy also raises to some extent. 

\begin{table}[htbp]
  \centering
  \caption{Different loss function applications in adversarial training. \textit{Inner, outer}, or \textit{both} denote to replace Cross-Entropy loss (CE) in the inner maximization of $\mathcal{L}_A$, outer minimization of $\mathcal{L}_T$, or both of them of Eqn.2 in the paper, respectively. A batch size of 128 is used here \textit{different} from the main paper, which does not affect the relative comparison among them.}
  \small
    \begin{tabular}{m{3.6cm}<{\centering}|m{0.75cm}<{\centering}|ccc}
    \toprule
    \textbf{Method} & \textbf{Apply} & \textbf{Clean} & \textbf{PGD} & \textbf{AA} \\
    \midrule
    CE & both  & 62.29  & 28.14  & 26.78  \\
    \midrule
    \multirow{3}[2]{*}{Class-aware margin~\cite{cao2019ldam}} & inner & 61.27 & 28.22 & 28.23 \\
          & outer & 60.70 & 28.04 & 26.75 \\
          & both  & 60.79 & 28.13 & 26.97 \\
    \midrule
    \multirow{3}[2]{*}{Re-weighting~\cite{cui2019cb}} & inner & 66.77 & 22.15 & 21.07 \\
          & outer & 62.76 & 32.76 & 27.77 \\
          & both  & 62.78 & 33.32 & 27.94 \\
    \midrule
    \multirow{3}[2]{*}{Class-aware temperature~\cite{ye2020identifying}} & inner & 63.98 & 26.89 & 25.96 \\
          & outer & 72.93 & 30.71 & 29.45 \\
          & both  & 72.70 & 28.26 & 27.21 \\
    \midrule
    \multirow{3}[2]{*}{Class-aware bias~\cite{menon2020logit, ren2020balanced-softmax}} & inner & 64.09 & 27.27 & 27.31 \\
          & outer & 71.33 & 29.25 & 27.82 \\
          & both  & 73.00 & 29.67 & 28.28 \\
    \bottomrule
    \end{tabular}%
    \vspace{-20pt}
  \label{tab:loss_function}%
\end{table}%



%
\subsection{Effect of PGD Steps during Training}
\label{supp:sec:pgd_step}
We use an iteration number of $5$ with the step size set as $2/255$, approximately $0.0078$, for the adversarial training procedure. We adopt this setting for an acceptable balancing of natural and robust accuracy of the baseline.
We study the effect of PGD iterations and step sizes in Table~\ref{tab:pgd_steps}. As the iteration number increases, the natural accuracy is improved along with the decline of robust accuracy. Especially for CIFAR-10-LT that when we change from 5 steps to 7 steps, there is a sharp decrease in clean accuracy.
As a result, we choose a 5-step PGD for the adversarial training framework in the paper.
\begin{table}[htbp]
  \centering
  \caption{Effect of different iteration numbers and step size in the inner maximum of the adversarial training procedure.}
  \small
    \begin{tabular}{c|c|cc||cc}
    \toprule
    \multicolumn{2}{c|}{\textbf{Adversarial Training}} & \multicolumn{2}{c||}{\textbf{CIFAR-10-LT}} & \multicolumn{2}{c}{\textbf{CIFAR-100-LT}} \\
    \midrule
    \textbf{Iterations} & \textbf{Step size} & \textbf{Clean} & \textbf{PGD} & \textbf{Clean} & \textbf{PGD} \\
    \midrule
    1     & 0.031 & 64.94 & 25.39 & 47.96 & 14.23 \\
    3     & 0.010 & 64.03 & 26.44 & 47.33 & 15.32 \\
    5     & 0.0078 & 62.29 & 28.14 & 46.16 & 15.91 \\
    7     & 0.0078 & 58.92 & 29.70 & 45.23 & 16.82 \\
    10    & 0.0078 & 57.61 & 29.27 & 45.31 & 17.40 \\
    \bottomrule
    \end{tabular}%
  \label{tab:pgd_steps}%
\end{table}%

\subsection{Intrinsic Properties among Classes}
\label{supp:sec:intrinsic}
%
Apart from the distribution of sample numbers, different intrinsic properties and the confusion cross categories are also non-negligible factors that lead to varying performance among classes.
As could be seen in Fig.1, when trained on balanced CIFAR-10, the difference in $A_{nat}$ is relatively minor, while it reveals the disparity of difficulty and vulnerability among classes, leading to a significant variance in $A_{rob}$. Specifically, Class 2, 3, and 4 demonstrate significantly lower robust accuracy compared with others.

\begin{figure}[htbp]
	\centering
	\vspace{-10pt}
	\includegraphics[width=1.0\linewidth]{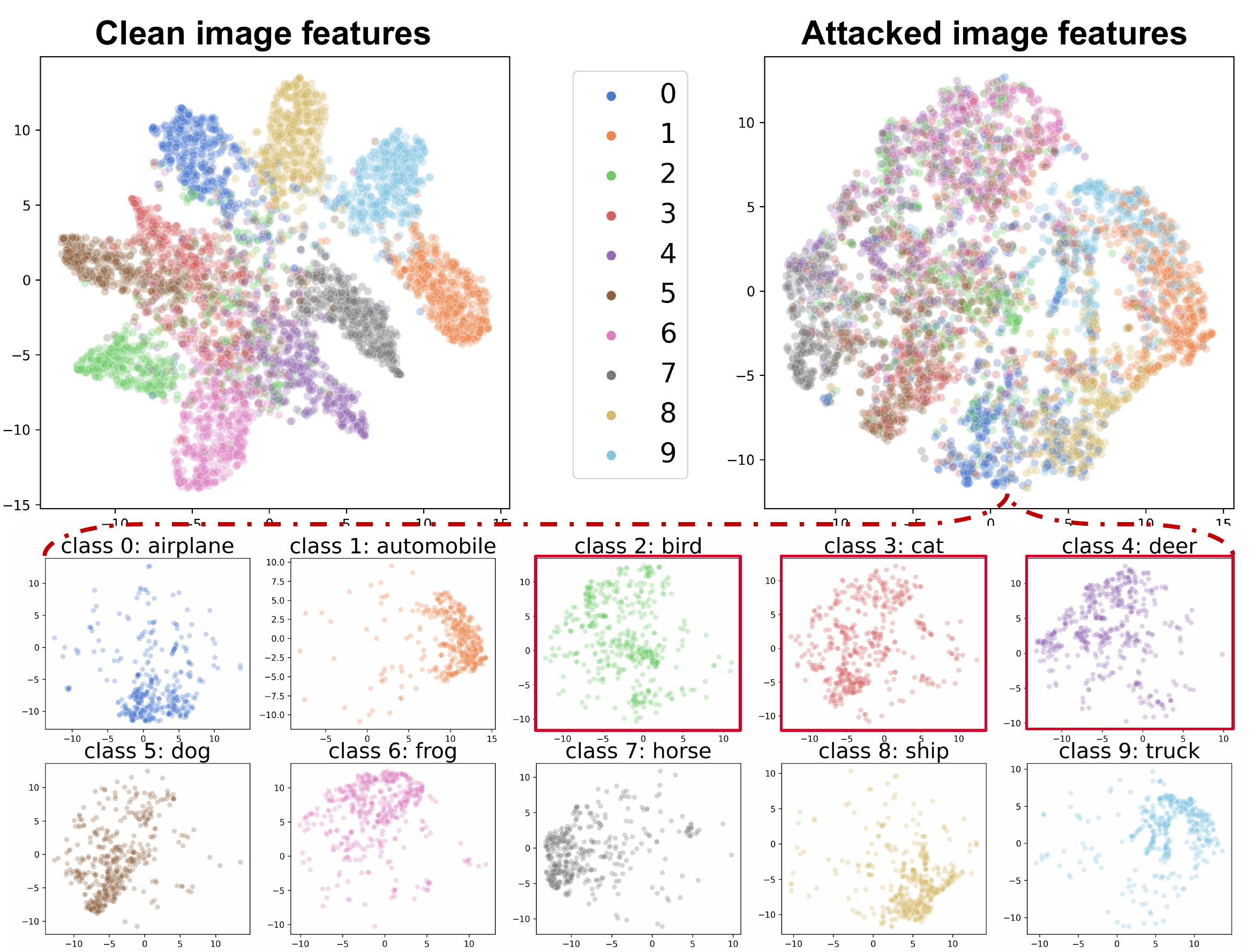}
	\caption{
	Latent space visualization before and after the attack. 
	}
	\vspace{-6pt}
	\label{fig:tsne}
\end{figure}

To study this phenomenon, we train a network on the balanced CIFAR-10 and visualize the latent space via t-SNE in Fig.~\ref{fig:tsne}. It shows that the classes with lower $A_{nat}$, such as Class 3, obviously have less concentrated and partially overlapped distributions, making them easier to be attacked. It can also be observed that Class 2, 3, and 4 have clearly more dispersed distributions under the attack, which is consistent with their low $A_{rob}$.
While under the long-tailed distribution, Class 3 benefits from the advantage of sample numbers over Class 4-9. Therefore, its accuracy becomes even higher than the original uniform distribution with the help of the induced prediction bias. 
A joint analysis of the effect by both intrinsic properties and the distribution of sample numbers among classes would be an interesting direction in the future.

\subsection{Experiments on ImageNet-LT}
\label{supp:sec:imagenet-lt}
We also evaluate our method on the more complicated ImageNet-LT~\cite{liu2019largescale} to encourage the exploration of real-world robustness. Due to the high resolution and large data scale, we adopt the standard single-step adversarial training (FGSM) and Fast adversarial training~\cite{wong2019fast}. We use ResNet-50 as the backbone with $\epsilon=2/255$ and $4/255$ following~\cite{shafahi2019free,wong2019fast}. The preliminary results are shown in Table~\ref{tab:imagenet-lt}. 

\begin{table}[h]
  \vspace{-3pt}
  \centering
  \small
  \caption{Adversarial robustness results on ImageNet-LT.}
    \begin{tabular}{l|c|ccc}
    \toprule
    Method & $\epsilon$ & CLEAN & FGSM  & PGD-20 \\
    \midrule
    FAST-AT & \multirow{2}[2]{*}{2 / 255} & 11.36 & 8.23  & 7.16 \\
    FAST-Our &       & \textbf{15.45} & \textbf{11.51} & \textbf{10.31} \\
    \midrule
    FGSM-AT & \multirow{2}[2]{*}{2 / 255} & 25.64 & 15.32  & 14.59 \\
    FGSM-Our &       & \textbf{30.02} & \textbf{18.50} & \textbf{17.67} \\
    \midrule
    FAST-AT & \multirow{2}[2]{*}{4 / 255} & 7.20  & 4.52  & 3.76 \\
    FAST-Our &       & \textbf{10.76} & \textbf{7.28}  & \textbf{6.13} \\
    \midrule
    FGSM-AT & \multirow{2}[2]{*}{4 / 255} & 21.94 & 10.88  & 9.45 \\
    FGSM-Our &       & \textbf{25.88} & \textbf{13.49} & \textbf{11.87} \\
    \bottomrule
    \end{tabular}%
    \vspace{-3pt}
  \label{tab:imagenet-lt}%
\end{table}%

Experimental results validate the effectiveness of our approach over the baseline. 
The relatively lower performance on ImageNet-LT compared to CIFAR also indicates that adversarial defense on the 1000-class ImageNet-LT is a more challenging problem, which is worth further exploration by the community.

\section{Adversarial Attacks}
\label{supp:sec:attacks}

\noindent\textbf{Fast Gradient Sign Method (FGSM)~\cite{goodfellow2015FGSM}} is a single-step attack that generates adversarial examples through a permutation along the gradient of the loss function with respect to the clean image as:
\begin{equation}
    x^{adv} = x + \epsilon \cdot sign(\nabla_x\mathcal{L}_{CE}(x_t^{adv},y)).
    \label{supp:eq:FGSM}
\end{equation}

\noindent\textbf{Projected Gradient Descent (PGD)~\cite{madry2018towards}} starts from an initialization point that is uniformly sampled from the allowed $\epsilon-ball$ centered at the clean image, and it extends FGSM by iteratively applying multiple small steps of permutation updating with respect to the current gradient as:
\begin{equation}
   x^{adv}_{t+1} = clip_{x, \epsilon}(x_t^{adv} + \eta \cdot sign(\nabla_x\mathcal{L}_{CE}(x_t^{adv},y))).
    \label{supp:eq:PGD}
\end{equation}

\noindent\textbf{Momentum Iterative gradient-based Methods (MIM) }~\cite{dong2018MIM} integrates the momentum into BIM with a decay factor $\mu$,
\begin{equation}
   g_{t+1} = \mu \cdot g_t + \frac{\nabla_x\mathcal{L}_{CE}(x_t^{adv},y)}{\left\Vert\nabla_x\mathcal{L}_{CE}(x_t^{adv},y)\right\Vert_1},
    \label{supp:eq:MIM1}
\end{equation}
and the permuted image is updated by:
\begin{equation}
   x^{adv}_{t+1} = clip_{x, \epsilon}(x_t^{adv} + \eta \cdot sign(g_{t+1}))).
    \label{supp:eq:MIM2}
\end{equation}

\noindent\textbf{Carlini \& Wagner (C\&W)~\cite{carlini2017CW}} is another powerful attack based on optimization, where an auxiliary variable $\omega$ is induced and  an adversarial example constrained by $l_2$ norm is represented by $x' = \frac{1}{2}(\tanh{\omega}+1)$. It can be optimized by:
\begin{equation}
  \argmin_{\omega}\{c \cdot f(x') + \left\Vert x'-x \right\Vert^2_2 \},
\label{supp:eq:CW1}
\end{equation}
where
\begin{equation}
    f(x') = \max(\max_{i\neq y}Z(x') - Z(x')_y , -\kappa),
\label{supp:eq:CW2}
\end{equation}
and here $\kappa$ controls the confidence of the adversarial examples.
It can also be extended to other $l_p$ threat model by solving $c \cdot f(x+\delta)+||\delta||_p$ in an iterative manner.

\noindent\textbf{Auto Attack~\cite{croce2020AA}} is a combination of multiple attacks that forms a parameter-free and computationally affordable ensemble of attacks to evaluate adversarial robustness. The standard attacks includes four selected attacks: \textit{$APGD_{CE}$}, targeted version of \textit{APGD-DLR} and \textit{FAB}, and \textit{Square Attack}.
Here we use the first two in our evaluation, because since the attack is applied in a curriculum manner, we empirically observe that after targeted \textit{APGD-DLR}, basically few adversarial examples are further explored by the last two attacks. So the change in the tested results of robust accuracy is quite small while the evaluation time can be significantly shortened.
{\small
\bibliographystyle{ieee_fullname}
\bibliography{egbib}
}

\end{document}